%% file: main.tex
\documentclass[12pt]{article}

\usepackage{arxiv}

\usepackage[utf8]{inputenc} 
\usepackage[T1]{fontenc}    
\usepackage{hyperref}       
\usepackage{url}            
\usepackage{booktabs}       
\usepackage{amsfonts}       
\usepackage{nicefrac}       
\usepackage{microtype}      
\usepackage{lipsum}

\usepackage[small]{caption}
\usepackage{amsmath,amssymb}
\usepackage{booktabs}
\usepackage{siunitx}
\usepackage{multirow}
\usepackage{xspace}
\usepackage{graphicx,lipsum,wrapfig,floatflt}  
\usepackage{subfigure}
\graphicspath{{./figures/}}

\include{shortcut}

\title{Partially Observable Planning and Learning for Systems with Non-Uniform Dynamics}

\author{
  Nicholas Collins \\
  School of ITEE\\
  The University of Queensland\\
  \texttt{nicholas.collins2@uq.net.au} \\
   \And
 Hanna Kurniawati \\
  Research School of Computer Science\\
  Australian National University\\
  \texttt{hanna.kurniawati@anu.edu.au} \\
}

\begin{document}
\maketitle

\begin{abstract}
\input{abstract.tex}
\end{abstract}


\vspace{-6pt}
\section{Introduction}\vspace{-6pt}
\input{intro.tex}

\vspace{-6pt}
\section{Background}
\input{background.tex}

\vspace{-9pt}
\section{\nop}\vspace{-9pt}
\input{approach_overview.tex}
\input{approach_full_detail.tex}

\vspace{-9pt}
\section{Computational Complexity}\vspace{-6pt}
\input{complexity.tex}
\vspace{-6pt}
\section{Experiments}\vspace{-3pt}
\label{experiments}
\input{results.tex}

\vspace{-9pt}
\section{Conclusion and Future Work}\vspace{-9pt}
\input{conclusion.tex}
\bibliographystyle{unsrt}  
\bibliography{references}  

\clearpage
\appendix
\section{Supplementary 1: Learned Transition Models}
\input{appendix_t_models.tex}

\end{document}

%% file: shortcut.tex
\newcommand{\cref}[1]{~\cite{#1}}

\newcommand{\fref}[1]{Figure~\ref{#1}}
\newcommand{\tref}[1]{Table~\ref{#1}}

\newcounter{comment}

\newcommand{\nop}{\textrm{TransNet}\xspace}

\newcommand{\pomdpTuple}{\textrm{$\langle S, A, O, T, Z, R, \gamma \rangle$}\xspace}

\newcommand{\bel}{\textrm{$b$}\xspace}

\newcommand{\sSpace}{\textrm{$S$}\xspace}

\newcommand{\actSpace}{\textrm{$A$}\xspace}

\newcommand{\obsSpace}{\textrm{$O$}\xspace}
\newcommand{\obs}{\textrm{$o$}\xspace}

%% file: abstract.tex
We propose a neural network architecture, called {\em \nop}, that combines planning and model learning for solving Partially Observable Markov Decision Processes (POMDPs) with non-uniform system dynamics. The past decade has seen a substantial advancement in solving POMDP problems. However, constructing a suitable POMDP model remains difficult. Recently, neural network architectures have been proposed to alleviate the difficulty in acquiring such models. Although the results are promising, existing architectures restrict the type of system dynamics that can be learned ---that is, system dynamics must be the same in all parts of the state space. \nop relaxes such a restriction. Key to this relaxation is a novel neural network module that classifies the state space into classes and then learns the system dynamics of the different classes. \nop uses this module together with the overall architecture of QMDP-Net\cite{karkus2017qmdpnet} to allow solving POMDPs that have more expressive dynamic models, while maintaining efficient data requirement. Its evaluation on typical benchmarks in robot navigation with initially unknown system and environment models indicates that \nop substantially out-performs the quality of the generated policies and learning efficiency of the state-of-the-art method QMDP-Net.

%% file: intro.tex
Sequential decision making under uncertainty is both critical and challenging. Partially Observable Markov Decision Processes (POMDPs) are the general and systematic frameworks for computing such decision making problems. Although finding the optimal strategies under the  POMDP framework is computationally intractable, advances have been made in computing approximately optimal strategies\cite{Kur08:Sarsop,Pin03:Point,Sil10:Monte,Som13:Despot}. In fact, we now have algorithms that can find near optimal strategies within reasonable time and have been applied to solve various realistic robotics problems (e.g., \cite{Bou17:Belief, Chen18:Human, Hoe19:POMDP}). 

With POMDP solving becoming practical and POMDPs starting to be used in practice, the problem of generating a good POMDP model for a given problem becomes increasingly important. A POMDP model is defined by six components: The states the system might be in, the actions it can perform, the observations it can perceive, system dynamics which represent uncertainty in the effect of actions, an observation function which represents uncertainty in sensing, and a reward function from which the objective function is derived. While the first three components are easy to define, the last three are more difficult due to uncertainty in the system and imperfect or even non-existent measurements to assess them. 

Many machine learning techniques have been proposed to alleviate the above difficulty\cite{ghavamzadeh2015bayesian,arulkumaran2017deep}. They can be divided into two broad classes. First is model-free, where the system learns a direct mapping from environment information to strategies, bypassing model generation. Second is  model-based, where the system first learns the model, and strategies are generated by applying model-based planning techniques to the learned model. 

Recently, deep neural networks have been proposed to combine model-free learning and model-based planning\cite{karkus2017qmdpnet,vin}. These works learn a direct mapping from environment information to strategies. However, internally these methods learn a POMDP model (or in the case of \cite{vin}, an MDP model ---a sub-class of POMDP where states are fully observable) and use a planning module, embedded inside the neural network, to generate the strategy. The objective of the model learning component here is not to generate the most accurate model, but rather to generate a \emph{useful} approximate model that will maximise policy performance when used together with the embedded planning algorithm. The results have been promising. However, they assume the system dynamics ---aka, transition function--- to be the same everywhere, regardless of the geometry of the underlying environment, which often limits the expressiveness of the model and restricts the effectiveness of planning.

To relax the above assumption, we propose a novel neural network architecture, called \nop. Key to \nop is a differentiable neural network module that learns non-uniform transition dynamics efficiently by assuming that states with similar local characteristics have similar dynamics. This module divides the state space into classes, where each class corresponds to a unique transition function. The transition probabilities for each class are then represented by the channels of a kernel in a convolution layer. This technique allows distinct transition dynamics to be applied to states with different local characteristics  while still allowing the use of existing efficient implementations of convolutional network layers. \nop uses this novel neural network module together with the overall architecture of state-of-the-art QMDP-Net to solve POMDPs with a priori unknown model and non-uniform transition dynamics.

Simulation experiments on various navigation benchmark problems with and without dynamic elements indicate that compared to QMDP-Net, \nop requires substantially less training time and data to produce policies with better quality: In some cases, TransNet uses less than 20\% of the training data used by QMDP-Net to generate policies with similar quality. Our results also indicate that \nop provides substantially better generalization capability than QMDP-Net.

%% file: background.tex
\vspace{-3pt}\subsection{POMDP Framework}\vspace{-3pt}


Formally, a POMDP\cite{kaelbling1998planning} is described by a 7-tuple \pomdpTuple, where \sSpace is the set of {\em states}, \actSpace is the set of {\em actions}, and \obsSpace is the set of {\em observations}.  At each step, the agent is in some hidden state $s\in \sSpace$, takes an action $a\in \actSpace$, and moves from $s$ to another state $s' \in \sSpace$ according to a conditional probability distribution $T(s, a, s') = P(s' | s, a)$, called the transition probability. The current state $s'$ is then partially revealed via an observation $o$ drawn from a conditional probability distribution $Z(s', a, o) = P(o | s', a)$ that represents uncertainty in sensing. After each step, the agent receives a reward $R(s,a)$, if it takes action $a$ from state $s$.  

Due to the uncertainty in the effect of action and in sensing, the agent never knows its exact state. Instead, it maintains an estimate of its current state in the form of a {\em belief} $b$, which is a probability distribution function over $\sSpace$.  At the end of each step, the agent updates its belief in a Bayesian manner, based on the belief at the beginning of the step along with the action and observation that have just been performed and perceived in this step.

The objective of a POMDP agent is to maximize its expected total reward ---called {\em value function}---, by following the best policy/strategy at each time step. A {\em policy} is a mapping from beliefs to actions. Each policy $\pi$ induces a value function $V_{\pi}(\bel)$ for any $\bel \in \mathbb{B}$, which is computed as: 
 \begin{equation}
V_{\pi}(b) = \sum_{s \in S} R(s, \pi(b)) \; b(s)  + \gamma \sum_{o \in O}P(o | b', \pi(b)) \; V_{\pi}(b') 
\label{e:valComplete}
\end{equation}
The notation $b'$ represents the new belief of the agent after it performs action $\pi(b) \in \actSpace$  and perceives observation \obs afterwards. It is computed as $b'(s') = \eta \sum_{o \in O} \; \sum_{s \in S} Z(s', \pi(b), o) \; T(s, \pi(b), s') \; b(s) $ ($\eta$ is a normalizing factor). When the planning horizon is infinite, to ensure the problem is well defined, rewards at subsequent time steps are discounted by a constant factor $\gamma \in (0, 1)$. The best policy $\pi^*$ is one that maximizes the value function at each belief \bel. 


\vspace{-3pt}\subsection{Related Work}\vspace{-3pt}


Recently, there has been a growing body of works that apply model-free deep learning to solve large scale POMDPs when the model is not fully known. For instance,  \cite{deep_learning_pomdp}  implemented a variation of DQN \cite{deep_learning_2} which replaces the final fully connected layer with a recurrent LSTM layer to solve partially observable variants of Atari games. The work in \cite{dl_pomdp_2} applied convolutional neural networks with multiple recurrent layers for the task of navigating within a partially observable maze environment. The learned policy is able to generalise to different goal positions within the learned maze, but not to previously unseen maze environments.

More recently, greater success have been achieved with methods that embed specific computational structures representing a model and algorithm within a neural network and training the network end-to-end, a hybrid approach which has the potential to combine the benefits of both model-based and model-free methods. For instance, \cite{vin} developed a differentiable approximation of value iteration embedded within a convolutional neural network to solve fully observable Markov Decision Process (MDP) problems in discrete space, while \cite{okada2017path} implemented a network with specific embedded computational structures to address the problem of path integral optimal control with continuous state and action spaces. These works focus only on cases where the underlying state is fully observable.

By combining the ideas in the above work with recent work on embedding Bayesian filters in deep neural networks\cite{hist_filter,backprop_kf,karkus2018particle}, one can develop neural network architectures that combine model-free learning and model-based planning for POMDPs. For instance, 
\cite{rcnn_qmdp} implemented a network which implements an approximate POMDP algorithm based on  $Q_{MDP}$ \cite{qmdp_original} by combining an embedded value iteration module with an embedded Bayesian filter. Modules are trained separately, with a focus on learning transition and reward models over directly learning a policy. 

More recently, \cite{karkus2017qmdpnet} developed QMDP-Net, which implements a $Q_{MDP}$ approximate POMDP algorithm to predict approximately optimal policies for tasks in a parameterised domain of environments. Policies are learned end-to-end, focusing on learning an ``incorrect but useful'' model which learns to optimise policy performance over model accuracy. However, the embedded model is restricted to using a simple transition model which assumes all states have the same transition dynamics. The transition function is represented as a kernel whose depth is the same as the size of the action space. The same learned kernel is applied to each state in the state space. This representation of the transition function enables the dynamics learned for one state to be generalised to other states, reducing the amount of training data needed to learn transition dynamics for all states. But as a result, QMDP-Net cannot represent non-uniform transition dynamics. \nop relaxes this restriction, while maintaining data efficiency.

%% file: approach_overview.tex
\nop learns a near optimal policy end-to-end, for acting in a parameterized set of partially observable scenarios: ${\cal {W}}_{\Theta} = \{ W(\theta) | \theta \in \Theta \}$, where  $\Theta$ is the set of all possible parameter values. Each  parameter $\theta$ describes properties of the scenarios such as obstacle geometry and materials, position of static and dynamic obstacles, goal location, and initial belief distribution for a given task and environment.  \nop assumes that the problems of deciding how to act in the various scenarios in ${\cal{W}}_{\Theta}$ are defined as POMDPs with a common state space $\underline{S}$, action space $\underline{A}$ and observation space $\underline{O}$ but without a priori known transition, action, and observation functions. \nop learns the parameterized transition, observation, and reward functions suitable to generate a good policy for the set of scenarios in ${\cal {W}}_{\Theta}$, as it learns the policy.

Similar to QMDP-Net, \nop's overall structure is a Recurrent Neural Network with two interleaving blocks: Planning and Belief update. \fref{f:transnet}(a) illustrates this network. However unlike QMDP-Net, in each block, \nop uses a neural network module as described in the following subsection to learn a transition function that depends on both actions and local characteristics of the states, rather than actions alone, thereby allowing more expressive POMDP models to be learnt, while maintaining data efficiency.

\vspace{-6pt}\subsection{Learning Non-Uniform Transition Dynamics}\vspace{-3pt}


Key to \nop is a neural network module for learning the transition function of a set of parameterized POMDPs. Suppose $M(\theta) = (S, A, O, f_{T}(.|\theta), f_{Z}(.|\theta), f_{R}(.|\theta))$ is the POMDP problem that corresponds to a scenario $W(\theta) \in {\cal{W}}_{\Theta}$. To learn the transition function $f_{T}(.|\theta)$, the neural network module represents $f_{T}(.|\theta)$  by a combination of a learned kernel and a classification function. The classification function $c(s | \theta)$ is a surjection that maps each state $s \in S$ to a class index, based on features of the parameter $\theta$. The kernel represents the probability of transitioning into each of the states in a local neighbourhood for each action $a \in A$ and each class, with separate channels representing different pairs of actions and classes. The pair of action and class index is then used to select the suitable kernel channel.
	
Two properties are desirable for the classification function. First, states with similar local characteristics  should map to the same class, and states with highly dissimilar characteristics should map to different classes. Second, the number of distinct state classes produced by the classification should be large enough to represent the important distinct modes of the transition dynamics, but small enough to ensure that information learned about the dynamics of one state is allowed to generalise to as many other appropriate states as possible. 

To generate the above desirable properties, in this work, the classification function is constructed by selecting a number of features of the scenario parameter $\theta$ which correspond to the local  features. The classification function $c(s | \theta)$ then  maps each state $s \in S$ to a class index based on the combination of feature values of the state $s$. Let $N$ be the number of features and $f_{1}(s)$...$f_{N}(s)$ be the values of the features of state $s \in S$. The classification function $c(s)$ is:
$$c(s) = \sum_{1 \leq i \leq N} (M + 1)^{i - 1} f_{i}(s)$$ 
\noindent
where $M$ is the maximum value of any feature of any state $s \in S$.  The class index $c(s)$ of state $s$ indicates the transition model to use at  $s$. 
We denote the image of this function, which represents the set of possible state classes, as $C$.

As an example, in a 2D robot navigation problem where $\theta$ includes an image indicating whether each cell in the environment is an obstacle (represented by 1) or free space (represented by 0), the features can be selected to be the values of the cells to the {\em north}, {\em south}, {\em east} and {\em west} of the current cell based on this image. The function $c(s)$ is then defined as $f_{\textrm{North}}(s) + 2f_{\textrm{South}}(s) + 4f_{\textrm{East}}(s) + 8f_{\textrm{West}}(s)$. When a state $s$ is blocked by obstacles in only its north and east side for instance, $c(s) = 1 + 0 + 4 + 0 = 5$. Of course, the image does not have to be binary. It may also represent information such as terrain types, obstacle types with different elasticity, areas of the environment which are subject to change over time, etc., allowing this representation to generalise to a wide range of scenarios.

\begin{figure}
\vspace{-12pt}
\subfigure[Overall architecture.]{\includegraphics[width=7.5cm]{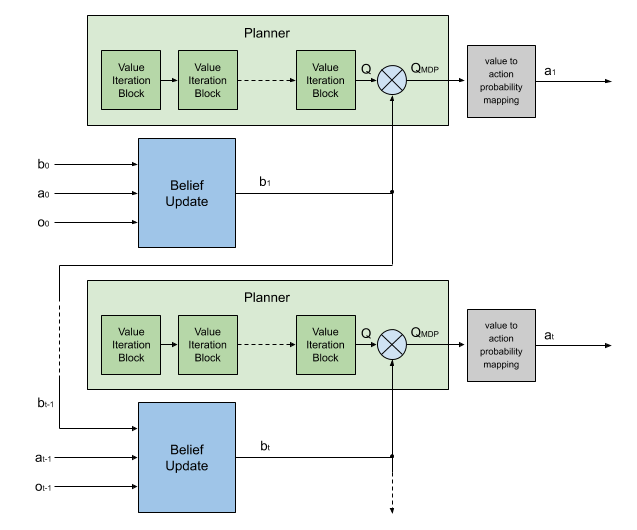}}
\subfigure[Key contribution of TransNet: A module that learns non-uniform transition function.]{
\includegraphics[width=8.5cm]{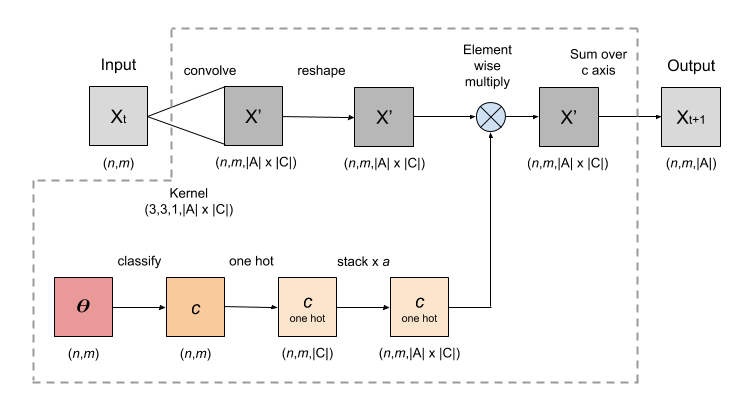}}
\caption{TransNet}
\label{f:transnet}
\vspace{-12pt}
\end{figure}

To avoid creating a bottleneck in the network, the classification function is implemented as a matrix operation in existing tensor libraries, allowing an image representing the state classification of every state in the state space to be computed efficiently for all states at once. Furthermore, a one-hot mapping is applied to the output of this function, which is then used to index into the channel corresponding to the local characteristics of each state using efficient matrix multiplication and summation operations. 
An illustration of \nop for a problem where the state space $S$ consists of two state variables, whose size is $n$ and $m$, respectively, is shown in \fref{f:transnet}(b).

The above manual selection of features and algorithmic classification could be replaced by an additional convolutional neural network, allowing important features which influence transition dynamics to be learned adaptively.

Now, one may question whether the same results can be achieved if we simply augment the state space with a state variable that indicates the class index of the original states, and then applying QMDP-Net to the augmented state space as is. The answer is  negative. Let's set aside the computational complexity issue caused by the substantially enlarged state space.  Since the transition function that QMDP-Net learns does not depend on states, this strategy of augmenting the state space will still ``force" a single transition function to be used for all states, including for states in different classes. In contrast, \nop learns multiple transition functions ---one function for each class. 


Note also that this module is general enough that it can be combined with any neural network architecture that embed POMDP/MDP planning with initially unknown transition function. However, \nop combines this module with QMDP-Net and embeds the module within every planning and belief update block. The following two subsections provide more details on this embedding.

%% file: approach_full_detail.tex
\vspace{-6pt}\subsection{Planning}\vspace{-3pt}

\begin{figure}[t]
\subfigure[A planning block of \nop]{\includegraphics[width=0.48\textwidth]{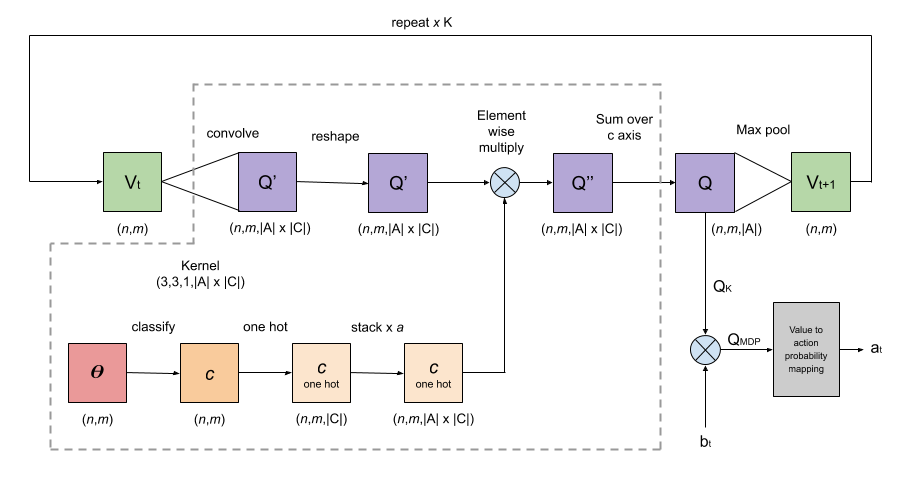}}
\hfill
\subfigure[A belief update block of \nop]{\includegraphics[width=0.48\textwidth]{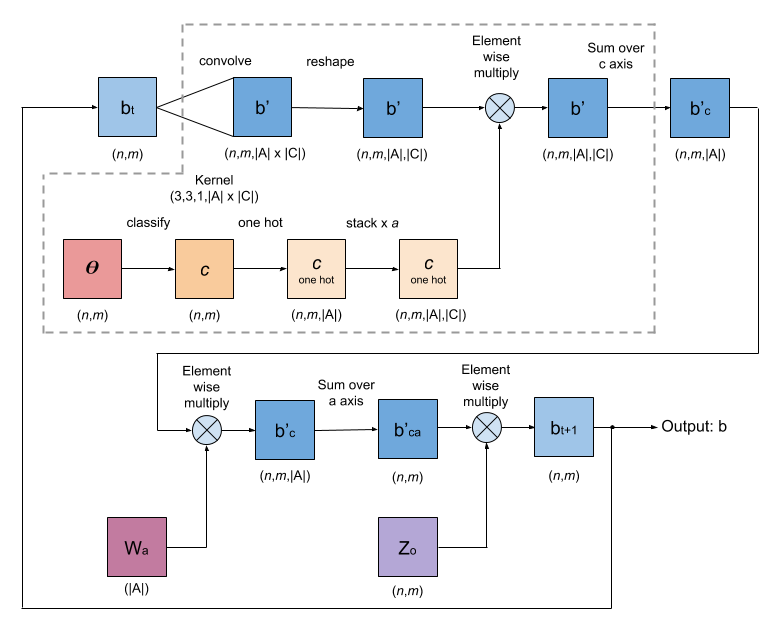}}
\caption{\nop architecture. Part of \nop that learns the transition function is marked by dashed-lines.}
\label{fig:planner}
\vspace{-12pt}
\end{figure} 
The planning component of \nop consists of a repeating block structure in which each block represents a single step of value iteration and blocks can be stacked to arbitrary depth to produce any desired planning horizon. Each block takes as input a value image $V_{t}(s|\theta)$, and produces as output updated values based on one additional planning step, $V_{t+1}(s|\theta)$, with the input to the first block, $V_{0}(s|\theta)$, taken from the prediction of the immediate reward associated with each $s \in S$ provided by $f_{R}$. 

\nop convolves the input with the neural network module for learning transition function. This module has one output channel for each pair $(a, c)$, where $a \in A$ and $c \in C$. The result of the convolution is a layer that represents the Q-values for each combination of state, action and  class index. Since for any scenario with parameter $\theta \in \Theta$, $c(s | \theta)$ is a surjection, we only need to select Q-values for the class that matches with $c(s | \theta)$. Therefore, the Q-values are multiplied with the one-hot representation of the state class image, before being summed over the axis corresponding to $c$. This has the effect of selecting the correct Q-values for the current $\theta$, and discarding all other invalid Q-values. These corrected Q-values are re-weighted by the belief. The maximum of these corrected Q-values over all $a \in A$ is then selected via a max-pooling layer to produce the updated value $V_{t+1}(s|\theta)$. The architecture of this block is illustrated in \fref{fig:planner}(a).

This implementation is a compromise, which sacrifices space complexity efficiency by computing and temporarily storing Q-values for classes which do not match $c(s|\theta)$ in order to facilitate the use of existing highly optimised implementations of convolutional network layers, without which training the network is infeasible.

\vspace{-6pt}\subsection{Belief Update}\vspace{-3pt}

A POMDP agent maintains a belief, which is updated at each time step using a Bayesian filter. To this end, \nop interleaves the planning block with the belief update block. The belief update block takes a prior belief $b_{t}$, action $a_{t}$ and observation $o_{t}$ as input, and produces the updated belief $b_{t+1}$ as output, which is stored as the prior belief for the next action selection. 

To compute $b_{t+1}$, \nop convolves $b_t$ with the neural network module for learning transition function. The resulting convolution is an image with one channel for each pair $(a, c)$, where $a \in A$ and $c \in C$, representing the updated probability of being in each state $s \in S$ for each combination of action and class index. The one-hot representation of the classes is used to select only the values for which class matches $c(s)$. A one-hot representation of the action $a_{t}$ applied at time $t$ is then used to select the values for which action matches $a_{t}$. The resulting belief represents the belief after accounting for the effect of the transition dynamics, notated as $b'$. 
A one-hot representation of the received observation $o_{t}$ is used to index into the observation model image predicted by $f_{Z}$ to produce an image indicating the predicted probability of receiving $o_{t}$ for each state $s \in S$. Finally this is used to weight $b'$ to produce the complete updated belief image, $b_{t+1}$. The architecture of a belief update block is shown in Figure \ref{fig:planner}(b).

%% file: complexity.tex
A key challenge in allowing non-uniform transition dynamics to be represented in a neural network structure is the complexity in terms of the number of trainable weights. By using classification, \nop significantly reduces this complexity. 

The number of trainable weights of \nop  is ${\mathcal{O}}(k^{2}|A||C|)$, where $|A|$ is the size of the action space, $|C|$ is the number of distinct state classifications, and $k^2$ is a constant that represents the size of the convolution filter for a single action and class. This constant is in general much smaller than the size of the state space. In the experiments presented in Section \ref{experiments}, $|C| \in [10, 100]$ were found to be practical and provide a good trade-off between cost and performance. 

The relatively low complexity would persist if the classification function based on manually selected features was replaced by an additional neural network. Suppose a network with a single convolutional layer with kernel width $k_{c}$ and a single fully connected layer with $h_{c}$ hidden neurons is used to assign a classification to each state $s \in S$. The number of weights in this network will be $k_{c}^{2}h + h|S|$. Combined with the learned transition probabilities for each class, the total number of weights used by the transition model is $k_{c}^{2}h + h_{c}|S| + k^{2}|C|$, giving complexity in terms of the number of trainable weights of ${\mathcal{O}}(h_{c}|S|)$, a reduction by a factor of approximately $k^{2}$.

%% file: results.tex
\vspace{-3pt}\subsection{Experimental Setup}\vspace{-3pt}

\begin{wrapfigure}{r}{7cm}
\vspace{-30pt}
\centering{\includegraphics[width=7cm]{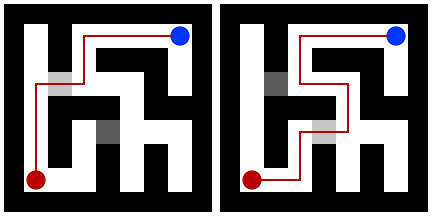}}
\caption{Example of a $9\times9$ dynamic maze environment in both possible gate states. Light grey represents an open gate, dark grey a closed gate. The agent must navigate from the red circle to the blue circle. The red line denotes the optimal trajectory.}
\label{fig:dynmaze}
\vspace{-12pt}
\end{wrapfigure}
To understand the practical performance of \nop, we compared \nop with state-of-the-art QMDP-Net. \nop's results are based on an implementation developed on top of the software released by the QMDP-Net authors, while QMDP-Net results are based on their released code. 

Both networks are trained via imitation learning using the same set of expert trajectories, with the expert trajectories generated by applying the $Q_{MDP}$ algorithm to manually constructed ground-truth POMDP models. Only trajectories where the expert was successful were included in the training set. The networks interact only with the expert trajectories and not with the ground-truth model. All hyper-parameters for both networks are set to match those used in the QMDP-Net experiments\cite{karkus2017qmdpnet}.

Training was conducted using CPU only on a machine with Intel Core-i7 7700 processor and 8GB RAM. We tested the networks on four domains: \vspace{-3pt}

\textbf{Gridworld Navigation:} A robot navigation problem in a general 2D grid setting with noisy state transitions and limited observations. The robot is given a map of obstacle positions, a specified goal location, and initial belief distribution. The robot must localise itself and navigate to the goal. At each time step, the robot selects a direction to move in, and receives a noisy observation indicating whether an obstacle is present in each of the ``north'', ``south'', ``east'' and ``west'' directions. The obstacle configuration is generated uniformly at random, with the constraint imposed that all non-obstacle cells are mutually reachable via some path.  \vspace{-3pt}

\textbf{Maze Navigation:} Similar to the gridworld navigation task, but with obstacle configuration generated using randomized Prim's algorithm. This results in expert trajectories typically being longer than in the general grid domain requiring longer term planning. This environment is also highly dependent on the planner's ability to identify dead-end passages.  \vspace{-3pt}

\textbf{Dynamic Maze Environments:} A navigation problem in a maze environment with structure that mutates during run-time in a way which qualitatively affects the optimum policy, designed to measure the robustness of a policy to dynamic environments. \\
A maze is initially constructed using randomized Prim's algorithm. The maze is divided into 2 partitions, with 2 cells from the border selected to be gates. At each time step, exactly one gate is open and the gates will swap from open to closed and vice versa with certain probability. The start and goal position are selected such that a gate swap will cause the optimum solution to be qualitatively changed. \fref{fig:dynmaze} illustrates an example. Two variations of this scenario are evaluated: \vspace{2pt}\\
	\textbf{V1:} The network is trained using only expert trajectories from the static maze navigation task. The environment image provided in $\theta$ shows only the positions of current free spaces and current obstacles, without special marking for open or closed gates. \vspace{2pt}\\
	\textbf{V2:} The network is trained using trajectories based on an expert which plans on a dynamic ground truth POMDP model, allowing the expert to decide whether to wait for a nearby closed gate to open. The environment image received by the agent denotes the position of the gate which is currently open. This may allow the agent to learn to intelligently decide whether to move or wait for the currently open gate to change. The closed gate is not represented in the image.  \vspace{-3pt}

\textbf{Large Scale Realistic Environments:} A navigation problem in realistic environments modelled on the LIDAR maps from the Robotics Data Set Repository \cite{Radish} with noisy actions and limited, unreliable observations. The network is trained on a set of randomly generated $10 \times 10$ stochastic grid environments, with the resulting policy then applied to the realistic environments, which have dimensions in the order of $100 \times 100$.

\vspace{-6pt}\subsection{Results and Discussion}\vspace{-3pt}

\input{tables/results_main.tex}

\tref{tab:res_main} presents comparisons on the success rate, average number of steps, and collision rate of executing the policies generated by QMDP-Net and by \nop. Training is conducted until convergence, but policies are outputted at a regular interval of 50 epochs. Training uses 10,000 different scenarios, comprising of 2,000 different environments and 5 different trajectories per environment. Policy evaluation is conducted on 500 different scenarios, comprising of 100 new environments and 5 different trajectories per environment.

The results indicate that \nop consistently produced substantially better policies than QMDP-Net and out-performs the training expert trajectories more consistently than QMDP-Net. The left side of \tref{tab:res_main} presents the results when training is run until convergence and comparison with the expert trajectory. In most cases, the number of epochs required to achieve convergence is lower in \nop than in QMDP-Net. Moreover, compared to QMDP-Net, \nop converges to policies with better quality. The right side of \tref{tab:res_main} presents the results where the training time are similar, giving slightly longer time to QMDP-Net. They indicate that although \nop requires more training time per epoch than QMDP-Net, \nop uses less time to generate policies with better quality. 

The results also demonstrate \nop is significantly more robust than QMDP-Net in dynamic environments. The success rate and collision rate of TransNet are not substantially degraded by the introduction of dynamic environment elements, and performance remains at or above the level of the QMDP expert trajectories.

\begin{table}[!htb]
  \centering
  \begin{minipage}{7.2cm}
  \caption{Comparison of the converged policy of \nop and QMDP-Net on Grid 10x10 S over different sizes of training set.}
  \resizebox{7cm}{!}{
    \begin{tabular}{ccrrrr}
    \toprule
    \textbf{Trajectories} & \textbf{Agent} & \multicolumn{1}{c}{\textbf{SR}} & \multicolumn{1}{c}{\textbf{TL}} & \multicolumn{1}{c}{\textbf{CR}} \\
    \midrule
    \multirow{2}[2]{*}{2000} & QMDP-Net  & 70.4 & 21.5  & 32.0 \\
          & TransNet & 98.2 & 15.3  & 11.2 \\
    \midrule
    \multirow{2}[2]{*}{10000} & QMDP-Net  & 95.0 & 15.1  & 13.9 \\
          & TransNet & 99.8 & 14.1  & 10.0 \\
    \midrule
    \multirow{2}[2]{*}{50000} & QMDP-Net  & 97.2 & 16.2  & 7.9 \\
          & TransNet & 99.2 & 15.4  & 6.8 \\
    \bottomrule
    \end{tabular}%
  }
  \label{tab:res_set_size}%
  \end{minipage}
  \hspace{0.5cm}
    \begin{minipage}{8cm}
  \caption{Comparison of the converged policy generated by \nop and QMDP-Net trained on Grid 10x10 D for (deterministic cases) and Grid 10x10 S for (for stochastic cases) and evaluated on large scale realistic environments derived from LIDAR datasets.}
  \resizebox{7.8cm}{!}{
    \begin{tabular}{ccrrr}
    \toprule
    \textbf{Domain} & \textbf{Agent} & \multicolumn{1}{c}{\textbf{SR}} & \multicolumn{1}{c}{\textbf{TL}} & \multicolumn{1}{c}{\textbf{CR}} \\
    \midrule
    \multirow{2}[2]{*}{Intel Lab 101x99 D} & QMDP-Net  & 40.0 & 100.0 & 6.6 \\
          & TransNet & 96.0 & 94.3  & 1.2 \\
    \midrule
    \multirow{2}[2]{*}{Intel Lab 101x99 S} & QMDP-Net  & 4.0 & 90.0  & 37.2 \\
          & TransNet & 68.0 & 129.2 & 3.7 \\
    \midrule
    \multirow{2}[2]{*}{Building 079 145x57 D} & QMDP-Net  & 56.0 & 70.8  & 22.5 \\
          & TransNet & 78.0 & 65.2  & 4.8 \\
    \midrule
    \multirow{2}[2]{*}{Building 079 145x57 S} & QMDP-Net  & 24.0 & 122.3 & 43.0 \\
          & TransNet & 52.0 & 107.5 & 7.9 \\
    \midrule
    \multirow{2}[2]{*}{Hospital 193x104 D} & QMDP-Net  & 14.0 & 85.1  & 28.6 \\
          & TransNet & 84.0 & 91.2  & 3.9 \\
    \midrule
    \multirow{2}[2]{*}{Hospital 193x104 S} & QMDP-Net  & 24.0 & 119.5 & 28.6 \\
          & TransNet & 52.0 & 193.3 & 4.2 \\
    \midrule
    \end{tabular}%
  }
  \label{tab:res_realistic}%
  \end{minipage}
\end{table}
Table \ref{tab:res_set_size} presents a comparison of the performance of \nop and QMDP-Net in a stochastic grid environment when trained on sets of expert trajectories of different sizes.

The results indicate \nop significantly reduces data requirements. \nop achieves a $98\%$ success rate after training with 2,000 scenarios. In contrast, QMDP-Net requires  50,000 scenarios to attain a comparable rate of success in this domain. The reduced data requirements enable \nop to be more practical for applications where acquiring training data is difficult or costly, such as when training data must be collected through interaction with a physical system.

Table \ref{tab:res_realistic} presents the generalization capability of \nop, compared to QMDP-Net. It compares the performance when networks trained on small artificially generated environments are  evaluated on large scale realistic environments: Intel Lab corresponds to the Intel Research Lab dataset, Building 079 corresponds to the Freiburg Building 079 dataset, and Hospital corresponds to the Freiburg University Hospital dataset.  To evaluate scenarios, we ran 25 trials per environment. In the work of \cite{karkus2017qmdpnet}, QMDP-Net was demonstrated to produce high rates of success on deterministic large scale environments when trained on expert trajectories in $30\times30$ random grids. Here, we trained both \nop and QMDP-Net on $10\times10$ random grids and evaluated in both deterministic and stochastic cases of realistic environments. 

The results indicate \nop substantially improves generalization capability. Local characteristics of states in the same class of problems (e.g., robot navigation in partially observed scenarios) tend to remain the same, even though the global complexity are totally different. Therefore, by learning separate transition functions based on local characteristics of the states, \nop can generate policies that generalized well.

We present the learned transition function for Grid10X10S in Supplementary-1. In summary, the transition learned is as expected.

%% file: tables/results_main.tex
\begin{table}[htbp]
\vspace{-12pt}
  \centering
  \caption{Performance comparison of \nop and QMDP-Net.  Expert is the QMDP algorithm. D indicates deterministic, S indicates stochastic. Epochs and time are  the number of epochs and training time taken to generate the policy. SR is the success rate (in \%) over all trials. TL is the average number of steps for successful trials. CR is the collision rate (in \%) over all steps.  Note that the number of steps required for completion is only directly comparable when success rates are similar.}
  \resizebox{1.0\textwidth}{!}{
    \begin{tabular}{cc|rrrr|rrrrr}
    \toprule
    & & \multicolumn{4}{c}{\textbf{Converged Policy}} & \multicolumn{5}{c}{\textbf{Policy after Similar Training Time}} \\
    \textbf{Domain} & \textbf{Agent} & \multicolumn{1}{c}{\textbf{Epochs}} & \multicolumn{1}{c}{\textbf{SR}} & \multicolumn{1}{c}{\textbf{TL}} & \multicolumn{1}{c}{\textbf{CR}} & \multicolumn{1}{c}{\textbf{Time (s)}} & \multicolumn{1}{c}{\textbf{Epochs}} & \multicolumn{1}{c}{\textbf{SR}} & \multicolumn{1}{c}{\textbf{TL}} & \multicolumn{1}{c}{\textbf{CR}}  \\
    \midrule
    \multirow{3}[2]{*}{Grid 10x10 D} & Expert & & 95.0 & 7.4  & 0.0  &  & & & &  \\
          & QMDP-net & 248  & 100.0 & 7.5   & 0.2 & 1,420 & 100 & 81.7 & 13.0 & 7.1 \\
          & TransNet & 328 & 100.0 & 7.5   & 0.0 & 1,310 & 50 & 89.8 & 8.6 & 7.5 \\
    \midrule
    \multirow{3}[2]{*}{Grid 10x10 S} & Expert & & 98.0 & 15.5  & 6.8 &  & & & &  \\
          & QMDP-net & 754  & 95.0 & 15.1  & 13.9 & 3,993 & 100 & 62.4 & 20.9 & 36.5 \\
          & TransNet & 543 & 99.8 & 14.1  & 10.0 & 3,092& 50 & 96.1 & 14.8 & 13.2 \\
    \midrule
    \multirow{3}[2]{*}{Maze 9x9 S} & Expert & & 88.4 & 15.5  & 10.5 &  & & & &  \\
          & QMDP-net  & 1,086 & 73.6 & 23.8  & 29.8 & 2,940 & 100 & 69.1 & 20.8 & 31.2 \\
          & TransNet & 837 & 97.8 & 15.6  & 15.9 & 2,257 & 50 & 83.0 & 18.7 & 23.6 \\
    \midrule
    \multirow{3}[2]{*}{Dynamic Maze V1 9x9 S} & Expert & & 85.2 & 23.3  & 13.1 &  & & & &  \\
          & QMDP-net & 1,565 & 71.0 & 25.8  & 33.9 & 2,982 & 100 & 62.1 & 24.7 & 32.8 \\
          & TransNet & 1,171 & 97.6 & 18.6  & 16.4 & 2,289 & 50 & 67.7 & 23.7 & 30.7 \\
    \midrule
    \multirow{3}[2]{*}{Dynamic Maze V2 9x9 S} & Expert & & 89.8 & 19.2  & 11.8 &  & & & &   \\
          & QMDP-net & 934  & 66.8 & 22.1  & 27.3 & 8,129 & 250 & 53.7 & 24.9 & 30.4 \\
          & TransNet & 1,122 & 87.6 & 19.1  & 15.5 & 7,902 & 50 & 63.5 & 22.6 & 20.8 \\
    \bottomrule
    \end{tabular}%
  }
  \label{tab:res_main}%
  \vspace{-12pt}
\end{table}%

%% file: conclusion.tex
\nop is a deep recurrent neural network for computing near optimal POMDP policies when the transition, observation, and reward functions are a priori unknown. The key novelty of \nop is a relatively simple neural network module that can learn non-uniform transition function efficiently. Experiments on navigation benchmarks indicate that \nop consistently out-performs state-of-the-art QMDP-Net. Moreover, results also indicate that \nop can generalize better and substantially reduce the amount of training data and time required to reach certain performance. 

This work suggests that a relatively simple neural network module can help embed more sophisticated models into deep neural networks, which then lead to substantial improvement for planning in stochastic domain. It is interesting to understand further how more sophisticated planning and learning components could help further scaling up of our capability in computing near optimal policies for decision making in stochastic domain.

%% file: appendix_t_models.tex
\begin{figure}[h]
\centering{\includegraphics[width=10cm]{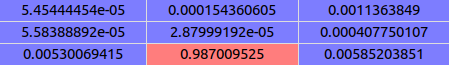}}
\caption{Planner transition model learned by QMDP-net for the 'Move South' action in the grid $10 \times 10$ S environment. Each square represents the probability of transitioning to each relative position (where the centre square represents the current state). Red indicates higher probability, blue indicates lower probability. Note that there is a high probability of transitioning to the south, and a low probability of transitioning in all other directions. These probabilities are applied in all situations, including when an obstacle is present to the south.}
\end{figure}

\begin{figure}[h]
\centering{\includegraphics[width=10cm]{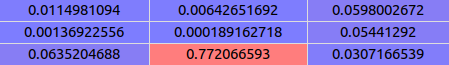}}
\caption{Planner transition model learned by TransNet for the 'Move South' action for the class where $c(s)=0$ (representing open space with no obstacle blockage in any direction) in the grid $10 \times 10$ S environment.}
\end{figure}

\begin{figure}[h]
\centering{\includegraphics[width=10cm]{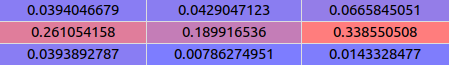}}
\caption{Planner transition model learned by TransNet for the 'Move South' action for the class where $c(s)=2$ (representing the case where an obstacle is present in the south direction with free space in all other directions) in the grid $10 \times 10$ S environment. Note that unlike in class 0, the probability of moving to the south is very low. This indicates that the model learned by the planner takes into account the fact that moving through an obstacle is not possible.}
\end{figure}